\documentclass{article}
\usepackage{arxiv}

\usepackage[utf8]{inputenc} 
\usepackage[T1]{fontenc}    

\usepackage{hyperref}       
\usepackage{url}            

\usepackage{booktabs}       
\usepackage{amsfonts}       
\usepackage{nicefrac}       
\usepackage{microtype}      
\usepackage{graphicx}
\usepackage{doi}
\usepackage[dvipsnames]{xcolor}
\usepackage{verbatim} 
\usepackage{amsfonts} 
\usepackage{caption}
\usepackage{tabularx}
\usepackage{multicol}
\usepackage{changepage}
\usepackage{natbib}

\title{Comparing credit risk estimates in the GEN-AI era}
\date{\today}
\author{
    \hspace{1mm}Nicola Lavecchia\\
    CRIF S.p.A.\\
    \texttt{n.lavecchia@crif.com} \\
        \And
        \hspace{1mm}Sid Fadanelli\\
    CRIF S.p.A.\\
    \texttt{s.fadanelli@crif.com} \\
    \And
    \hspace{1mm}Federico Ricciuti \\
    CRIF S.p.A. \\
    \texttt{f.ricciuti@crif.com} \\
    \And
    \hspace{1mm}Gennaro Aloe \\
    CRIF S.p.A. \\
    \texttt{g.aloe2@crif.com} \\
    \And
    \hspace{1mm}Enrico Bagli\\
    CRIF S.p.A.\\
    \texttt{e.bagli@crif.com} \\
    \And
    \hspace{1mm}Pietro Giuffrida \\
    CRIF S.p.A.\\
    \texttt{p.giuffrida@crif.com} \\
    \And
    \hspace{1mm}Daniele Vergari \\
    CRIF S.p.A.\\
    \texttt{d.vergari@crif.com} \\
}

\renewcommand{\headeright}{ }
\renewcommand{\undertitle}{ }
\renewcommand{\shorttitle}{ }

\begin{document}
\maketitle

\begin{abstract}
Generative AI technologies have demonstrated significant potential across diverse applications. This study provides a comparative analysis of credit score modeling techniques, contrasting traditional approaches with those leveraging generative AI. Our findings reveal that current generative AI models fall short of matching the performance of traditional methods, regardless of the integration strategy employed. These results highlight the limitations in the current capabilities of generative AI for credit risk scoring, emphasizing the need for further research and development before the possibility of applying generative AI for this specific task, or equivalent ones.
\end{abstract}

\section{Introduction}
Credit risk assessment is a critical process in the financial industry, involving the in-depth analysis of both structured and unstructured data to evaluate the likelihood of default. More specifically, credit risk scoring—an integral component of credit risk assessment—translates this comprehensive analysis into a quantitative measure that informs lending decisions. Historically, this task has been performed using traditional statistical methods like Logistic Regression model, which have demonstrated robust performance over time. \citep{article}.

In recent years, \textit{generative AI} (GenAI) models have witnessed remarkable advancements, generating widespread interest in their potential applications across various industries. In the financial sector, recent studies have highlighted the success of GenAI in tasks such as automating document reading and extracting key financial metrics from balance sheets and regulatory filings \citep{zou2024docbenchbenchmarkevaluatingllmbased}. While these applications underscore GenAI’s ability to process complex financial documents, they also raise questions about whether such strengths directly translate into improved performance in specialized task such as credit risk scoring.

In this regard, our study systematically contrasts the promising capabilities of GenAI with the time-tested robustness and interpretability of traditional credit risk models, which continue to set the benchmark in high-stakes, regulated decision-making contexts.

Although several studies have pointed in this direction \citep{YDeldjoo_2023, YLi_2024, DFeng_2024, GBabaei_2024}, a systematic comparison of traditional AI and GenAI models on a unified benchmark is still lacking. This gap leaves critical questions unanswered about the limitations of GenAI in credit risk scoring and its broader applicability within the field.

To address this issue, the present study aims to provide a comprehensive performance comparison between traditional models and their GenAI-based counterparts in credit risk assessment. Our preliminary findings suggest that traditional methods tend to outperform current GenAI implementations in terms of discrimination and calibration. 

\section{Procedure}
\subsection{Database}
The \textit{German Credit Risk} (GCR) dataset \citep{HHoffmann_1994} has become a widely recognized benchmark for evaluating credit scoring models, especially due to the rather limited performance of traditional evaluation methods when applied to it - also because of the small number of available observations. It comprises 1,000 records of loan applications, each characterized by 13 qualitative and 7 quantitative features. Approximately 70\% of the applications are classified as low risk, while the remaining 30\% are deemed high risk. 

For this study, the dataset has been split into a training set of 700 records and a testing set of 300 records. Initially, one single train-test split was performed using stratified sampling. As an additional cautionary measure, bootstrap sampling allowed to evaluate each of the models on 100 random splits of the database into 700-300 data point sections. 

From the 20 available features, only the 11 in the following table were selected for modeling. 
\begin{table}[h]
\centering
\begin{tabular}{l|l|l}
\textit{variable name}                  & \textit{type} & \textit{description} \\ \hline
\texttt{existing\_checking\_account}    & 4 values    & status of existing checking account \\ 
\texttt{duration}                       & numeric     & duration of the requested loan, in months \\ 
\texttt{credit\_history}                & 5 values    & indicates whether previous credits were repaid duly or not \\ 
\texttt{purpose}                        & 11 values   & expenditure of loan money, if granted (car, business, ...)\\ 
\texttt{credit\_amount}                 & numeric     & the amount of the requested loan, in DM \\ 
\texttt{savings/bonds}                  & 5 values    & consistence of existing savings in accounts or bonds \\ 
\texttt{installment\_rate}              & numeric     & installment rate, percentage of disposable income\\ 
\texttt{employment\_duration}           & 5 values    & the length of time the client has been employed\\
\texttt{installment\_plans}             & 3 values    & other installment plans\\
\texttt{housing}                        & 3 values    & applicant's living situation\\ 
\texttt{personal\_status}               & 5 values    & marital status\\ 
\texttt{gender}                         & 2 values    & anagraphic gender \\ 
\texttt{age}                            & numeric     & loan applicant age, in years\\ 
\texttt{foreign\_born}                  & 2 values    & indicates whether the loan applicant is foreign-born \\ 
\end{tabular}
\label{tab:feature-list}
\end{table}

The selection of these variables was effected by following two rationales. On the one hand, we considered statistical significance as retrieved by a sensitivity analysis carried on in a bidirectional stepwise regression fashion (binning predictors and calculating their weight of evidence values - see \citep{siddiqi2012credit} and \citep{saltelli2008global}). On the other hand, some variables have been selected to allow a comprehensive evaluation of fairness: this was the case for \texttt{age}, \texttt{gender}, and \texttt{foreign\_born}. By following both these criteria we have ensured that our models are based on the most predictive variables and allow us to assess potential biases in the credit risk assessment conducted by our models.

\subsection{Credit Risk Evaluators Based on GenAI}
In this study, credit risk evaluators based on GenAI were implemented as pipelines that process records from the GCR dataset into structured queries. These queries were then submitted via API to the \texttt{gpt-4o-2024-08-06} model provided by OpenAI, setting the temperature parameter to 0 to ensure deterministic and reproducible outputs. All queries to GenAI models have been set to return a binary assessment of credit risk, meaning that they are required to answer either "low" or "high" (see prompting details below). To derive continuous risk estimates from these generative outputs, the \texttt{logprobs} parameter was enabled via the API, and the probability assigned to the "low" class was inverted (i.e. computed as 1 minus the low probability) so that each model would ultimately provide us with a "creditor default probability" estimate. 

All evaluations have been performed within a few-shot learning approach, where a set of representative examples was included in the prompt to guide the model's predictions and enhance its ability to classify credit risk effectively. Two distinct prompting strategies were tested to evaluate the effectiveness of GenAI in this context.
\begin{itemize}
    \item \textbf{Structured Prompt Query} In this approach, each record from the dataset was encoded into a dictionary, where keys represented column names and values were presented in a human-interpretable format. The prompt was designed to set the context and provide detailed instructions to the model. Examples of records were included to guide the model's understanding of the task. The target record was structured in the same format as the examples but excluded the customer's risk assessment, which the model was tasked to predict.

    \item \textbf{Textual Prompt Query} This approach involved encoding records into textual descriptions that summarized the key attributes of each credit application. The structure of these descriptions was similar to that used in the structured prompt, with contextual instructions and examples provided. Like in the structured approach, the target record was presented without a risk assessment for the model to evaluate.
\end{itemize}

The structured prompt leverages a clear and standardized format, enabling the model to map relationships between variables effectively and reduce ambiguity. Conversely, the textual prompt takes advantage of the model's advanced understanding of natural language: even though unstructured data is uncommon in credit risk scoring, this approach enables to ascertain whether performances can be improved by adding semantic depth to the data. \\
\qquad \\


\begin{minipage}[t]{0.45\textwidth} 
\noindent\textbf{Structured Prompts} \\
\scriptsize 
\,\\
You are an expert in assessing the risk of providing a credit to the customers of a well known bank.
Your task is to analyze a fixed set of information related to a customer and associate to it a level of credit risk.
The credit risk can be defined in two classes: low or high.

Some examples of how you have to classify a customer are reported below:
\smallskip

\begin{verbatim}
{
    "Existing checking account": "between 0 and 200 $",
    "Duration": 6,
    "Credit history": "existing credits paid back 
                        duly till now",
    "Purpose": "radio/television",
    "Credit Amount": 2063,
    "Savings": "less than 100 $",
    "Installment Rate": 4,
    "Gender": "Male",
    "Age": 30,
    "Foreign Worker": true,
    "Employment Duration": "less than 1 year",
    "Personal Status": "married or widowed",
    "Installment Plans": "none",
    "Housing": "rent"
}

- Credit risk: low

.
.
.
.

\end{verbatim}

Consider them as indicative examples of how you should perform the task,
for example if a custom has characteristics similar to an example reported above, you should classify it with the same risk class.

Now you have to classify a new customer, below, the available information of the customer are reported:
\\

\begin{verbatim}
{
    "Existing checking account": "no checking account",
    "Duration": 48,
    "Credit history": "critical account / other 
            credits existing (not at this bank)",
    "Purpose": "radio/television",
    "Credit Amount": 3578,
    "Savings": "unknown / no savings account",
    "Installment Rate": 4,
    "Gender": "Male",
    "Age": 47,
    "Foreign Worker": true,
    "Employment Duration": "greater than 7 years",
    "Personal Status": "single",
    "Installment Plans": "none",
    "Housing": "own"
}

- Credit risk: 
    
\end{verbatim}
\smallskip
Provide a classification of the credit risk of the customer in one of the two risk categories: low or high.
Answer strictly with the risk category.

\end{minipage}
\hspace{\tabcolsep}\hspace{\tabcolsep}\vrule\hspace{\tabcolsep}\hspace{\tabcolsep}
\begin{minipage}[t]{0.5\textwidth} 
\noindent\textbf{Textual Prompts} \\
\scriptsize 
\,\\
You are an expert in assessing the risk of providing a credit to the customers of a well known bank.
Your task is to analyze a fixed set of information related to a customer and associate to it a level of credit risk.
The credit risk can be defined in two classes: low or high.

You have access to several information related to the customer contained in a description of his position.

Some examples of how you have to classify a customer are reported below:

\begin{verbatim}





A married or widowed, Male of 30-year-old customer, applied 
for a loan of 2063$ for radio/television. 
The customer has a checkingaccount status of 'between
0 and 200$' and savings of 'less than 100$'. 
The customer has been employed for less than 1 year.
The loan duration is 6 months, with a credit history described as 
'existing credits paid back duly till now'. The installment
rate is 4% of disposable income. Their available installment 
plans are 'none' and the housing situation is 'rent'.
The customeris a domestic worker.


- Credit risk: low

.
.
.
.

\end{verbatim}
Consider them as indicative examples of how you should perform the task,
for example if a custom has characteristics similar to an example reported above, you should classify it with the same risk class.

Now you have to classify a new customer, below, the available information of the customer are reported:
\begin{verbatim}




    

A single, Male of 47-year-old customer, applied for a loan of 3578$ 
for radio/television. The customer has a checking account status
of 'no checking account' and savings of 'unknown / no savings account'. 
The customer has been employed for greater than 7 years.
The loan duration is 48 months, with a credit history described 
as 'critical account / other credits existing (not at this bank)'.
The installment rate is 4% of disposable income. 
Their available installment plans are 'none' and the housing 
situation is 'own'.
The customer is a domestic worker.





- Credit risk: 

\end{verbatim}
\smallskip
Provide a classification of the credit risk of the customer in one of the two risk categories: low or high.
Answer strictly with the risk category.

\end{minipage}

\bigskip 


Also, we have considered various strategies to extract examples from the training dataset.
\begin{itemize}
    \item \textbf{Random Example Selection} In this approach, the model is provided with examples randomly selected from the training dataset. Random selection allows for the model not to "drown" in a sea of information, i.e. it should allow the GenAI to "learn" from the examples provided (with the set of examples remaining the same regardless of the target record being evaluated, ensuring consistency between evaluations). 

    \item \textbf{Neighbor Example Selection} This approach focuses on providing the model with only those examples which are the most similar to the target record, in the same spirit of the now-common retrieval-augmented generation procedures \citep{PLewis_2020}. In this case, a \textit{K-Nearest Neighbors} (KNN) algorithm selects $K$ records from the training dataset such that these are the most similar to the evaluation target. These records are then incorporated as examples into the few-shot learning prompt provided to the generative model. Note that the sample is generated by a balanced retrieval mechanism, in which neighbor distances get re-scaled inversely proportional to credit risk frequencies: neighbors with low credit risk, which are the most, will get their distances increased more than what is done for neighbors with high credit risk, which are less frequent. This is done in such a way that a sample with 50/50 proportion between high and low risk scores is obtained for any target point which is surrounded by a region in which low and high risk neighbors are in the same proportion as in the whole database. Had we not put in place this selection procedure, the GenAI models wouldn't have any way to access the fact that the baseline split is 30/70 in favor of low-risk neighbors. 
\end{itemize}
The use of both random-example queries and selected-example queries ensures a comprehensive evaluation of the model's behavior. Random examples provide a balanced and consistent representation of the dataset, allowing for a general assessment of the model's performance across diverse scenarios. In contrast, selected-example queries were specifically designed to ensure that generative and traditional models could be assessed on the same grounds. Indeed, while in training traditional models it is straightforward to employ thousands of data points, incorporating such a large number of examples in a few-shot prompt for a generative model would be both token-intensive and inefficient. In selected-example queries the limitation of generative models in terms of "training" examples is compensated by the selection of only the most representative examples. 

Thus, all GenAI evaluators developed in this study can be characterized by specifying three key parameters: the prompting strategy (structured or textual), the example selection procedure (random or selected), and the number of examples included in the prompt. Varying strategies and parameters allows us to assess in full the effectiveness of generative models in credit risk prediction.

\subsection{Benchmark Models}
As counterpart to the GenAI estimations, we developed credit risk evaluators to adopt as benchmark.

\begin{itemize}
    \item \textbf{Logistic Regression} Our logistic regression model was implemented following procedures from \citep{siddiqi2012credit} and software from the \texttt{scorecard} package \citep{xie2024scorecard}. 

    \item \textbf{Nearest-Neighbors Models} Our nearest-neighbor predictors are devised specifically to benchmark the efficiency of the process adopted to build the so-called "neighbor example selection" for GenAI models, The idea behind this approach is that if the example selection process devised is effective then a basic estimate of credit risk can be obtained by taking the average of the retrieved records' risk values. Averaging known risk values draws the bottom line for the selected-example GenAI estimators: any GenAI estimator falling behind a simple average cannot be deemed as successful.

\end{itemize}

\subsection{Evaluation Procedure}
Evaluation focused on two primary aspects: performance, defined as the ability to make generalizable predictions, and fairness, understood as the ability to treat equitably cases involving individuals with different sensitive attributes. In our analysis, we consider as sensitive variables the \texttt{age}, \texttt{gender}, and \texttt{foreign\_born} variables.

For performance evaluations we plotted \textit{receiver-operating-characteristic} (ROC) curves \citep{MJunge_2024} and calculated the following metrics.
\begin{itemize}
    \item \textit{Classification accuracy} i.e. the proportion of correctly classified instances once a threshold of 0.5 is applied across all predictions. Note that this metric, although classic, can be somewhat misleading in two aspects. First, the preliminary binning of risk evaluations implies that it depends fundamentally on the threshold parameter used to separate low from high scores. Second, unbalanced datasets generally make this into an inefficient or possibly misleading performance evaluator (see \citep{MGrandini_2020, DChicco_2017} for reference, and note that in our dataset the 30/70 split of target values implies that the perfectly random evaluator would have an accuracy of 0.7). 
    \item \textit{Gini coefficient} i.e. the normalization of the area under a ROC curve, an overall measure of discriminatory power which quantifies the inequality in risk scores between defaulters and non-defaulters. Credit risk models typically target a minimum Gini of 40\%, with high-performing models exceeding 60\%. 
    \item \textit{Kolmogorov-Smirnov statistics} i.e. the maximum difference between the cumulative distribution functions of defaulters and non-defaulters across risk scores, which can be visualized easily in a ROC diagram as it is proportional to the maximum distance curve and the main diagonal \citep{adeodato2016equivalencekolmogorovsmirnovroccurve}. This metric is particularly favored in the assessment of credit risk evaluators given that its simple definition allows for an immediate use in operational decision-making, such as setting cutoff values. 
    \item \textit{Logarithmic loss} which evaluates the predicted probabilities against the 0/1 benchmark by the cross-entropy of the two distributions relative to each other. The higher this value, the further our evaluations are from the benchmark.
    \item \textit{Brier score} which quantifies the accuracy of probability predictions as the mean squared error between predicted probabilities and actual outcomes. Also in this case the worst an evaluation is and the higher the value of this metric.
\end{itemize}
Fairness is assessed by two dedicated metrics. 
\begin{itemize}
    \item For a given sensitive feature, its \textit{equalized odds difference} (EOD) is determined by computing the \textit{true positive rates} (TPR) and \textit{false positive rates} (FPR) inside each group defined by the values of the sensitive feature, then taking the largest value among all TPR differences and all FPR differences - high values of EOD indicate that the model treats in a disparate way records with different values in the given sensitive feature. Since this fairness indicator is designed for a sensitive feature of categorical type, the \texttt{age} has been binned into six classes before calculating this metric. 
    \item The \textit{BRIO} risk metric evaluates the overall disparity in treatment that the model presents across the full set of sensitive features, with lower values indicating less biased predictors \citep{Coraglia2023BRIOxAlkemy, coraglia2024evaluatingaifairnesscredit} 
\end{itemize}


\begin{figure}[b]
\centering
{\includegraphics[width=.98\textwidth]{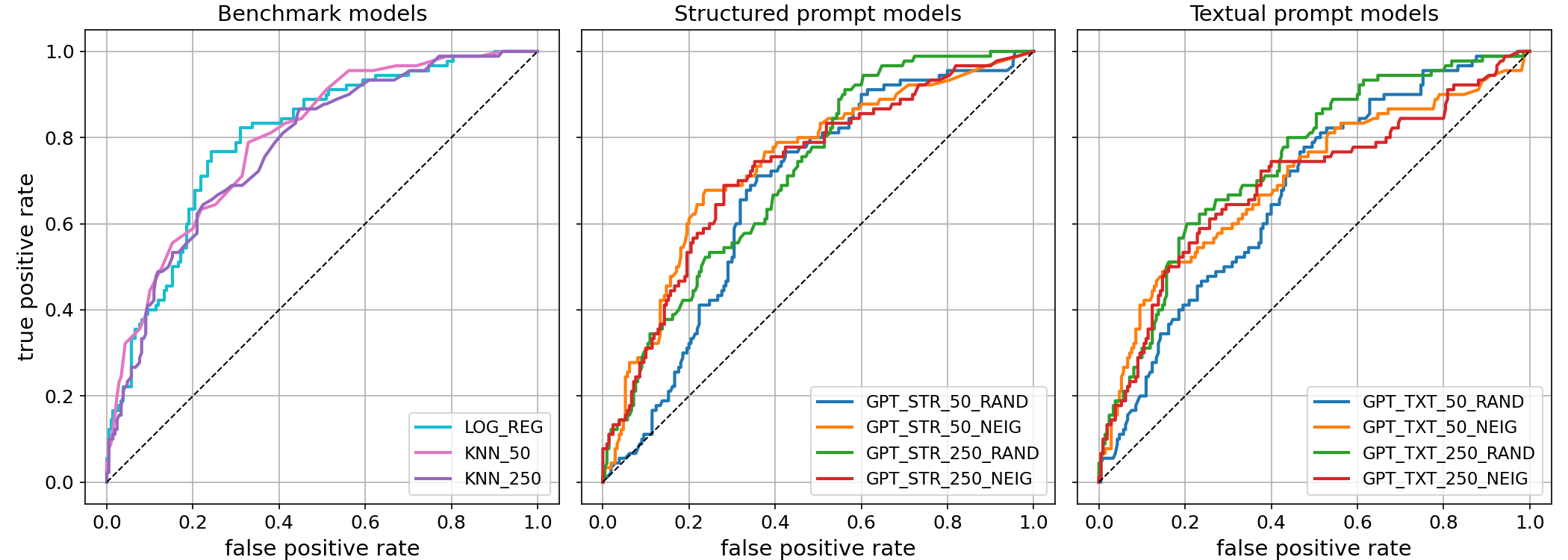}}
\caption{ROC curves for all the GenAI models. Description of all models in the text. }
\label{fig:roc_curves}
\end{figure}

\section{Results}
\subsection{Overview}

Our main investigation is based on eight GenAI models which explore all the combinations of our prompting strategies (structured or textual prompting, random or selected examples, sample size of 50 or 250 points) and three non-GenAI benchmark models (one retriever-only for each of the sample sizes we have selected, one last model based on logistic regression). Performance and fairness metrics evaluated on the test data from the first stratified sampling are reported in the in the table below, while the ROC curve of each model is plotted in figure \ref{fig:roc_curves} (graph on the left for benchmark models, graphs on the center and right for GenAI-based risk evaluators). Results obtained through bootstrap are not reported given their close alignment with those achieved using stratified sampling. 

\begin{table}[t]
\centering
\resizebox{\textwidth}{!}{%
\begin{tabular}{l|ccccc|cccc}
\textit{model} & \textit{Class. accuracy} & \textit{Gini coeff.} & \textit{K-S stat.} & \textit{Log. loss} & \textit{Brier score} & \textit{EOD: gender} & \textit{EOD: age} & \textit{EOD: foreign\_worker} & \textit{BRIO} \\ \hline\hline
LOG\_REG             & 0.737 & 0.596 & 0.524 & 0.490 & 0.163 & 0.119 & 0.231 & 0.455 & 0.131 \\ \hline
KNN\_50              & 0.763 & 0.593 & 0.467 & 0.510 & 0.168 & 0.119 & 0.401 & 0.568 & 0.142 \\ \hline
KNN\_250             & 0.737 & 0.552 & 0.435 & 0.549 & 0.182 & 0.112 & 0.157 & 0.807 & 0.142 \\ \hline
GPT\_TXT\_50\_RAND   & 0.533 & 0.350 & 0.302 & 1.510 & 0.394 & 0.016 & 0.308 & 0.172 & 0.064 \\ \hline
GPT\_TXT\_50\_NEIG   & 0.743 & 0.398 & 0.348 & 0.925 & 0.217 & 0.033 & 0.314 & 0.133 & 0.072 \\ \hline
GPT\_STR\_50\_RAND   & 0.583 & 0.346 & 0.354 & 1.482 & 0.348 & 0.074 & 0.344 & 0.232 & 0.101 \\ \hline
GPT\_STR\_50\_NEIG   & 0.730 & 0.477 & 0.444 & 1.201 & 0.215 & 0.057 & 0.214 & 0.443 & 0.088 \\ \hline
GPT\_TXT\_250\_RAND  & 0.723 & 0.488 & 0.395 & 0.702 & 0.219 & 0.010 & 0.316 & 0.386 & 0.096 \\ \hline
GPT\_TXT\_250\_NEIG  & 0.723 & 0.385 & 0.356 & 0.811 & 0.214 & 0.020 & 0.473 & 0.614 & 0.077 \\ \hline
GPT\_STR\_250\_RAND  & 0.647 & 0.432 & 0.349 & 0.823 & 0.259 & 0.260 & 0.417 & 0.177 & 0.131 \\ \hline
GPT\_STR\_250\_NEIG  & 0.717 & 0.451 & 0.408 & 0.874 & 0.221 & 0.092 & 0.429 & 0.170 & 0.074 \\ \hline
\end{tabular}%
} \\
\captionsetup{justification=raggedright}
\caption{Comparison of model performance}

\label{tab:benchmark-performance}
\end{table}

\subsection{Model performance}
In terms of performance, the most striking evidence is that non-GenAI models tend to beat GenAI evaluators across the board. Indeed, it is the logistic regression to come out as best model in all the four nonparametric indicators (the Gini coefficient, Kolmogorov-Smirnov statistics, logarithmic loss, and Brier score) with the KNN models scoring as the second and third best in all but one of these metrics. It is certainly significative that GenAI-based models do not evaluate risk at random when provided with random points (otherwise the ROC curves of all these models would fall on the diagonal) but nonetheless they fail rather spectacularly to score better than the simple average of risk values whenever examples are selected only among the closest data points. In other words, GenAI estimators are not yet at the level of their traditional counterparts, irrespective of prompting, sample selection strategy, and the number of data points provided to them as evidence - with the particularly dire remark that whenever examples are selected carefully, GenAI evaluators score worse than the simple average of risk values in the sample. 

Compared to the hiatus between GenAI and traditional models, other trends in performance are feeble at best. Yet, we can recognize that the logarithmic loss diminishes for all GenAI models as sample size is passed from 50 to 250 points, and that the Brier scores of GenAI models with randomly selected examples are always higher than their counterparts for models with neighbor example selection. These trends seem to hint to a slight improvement in performances as we move to a somewhat higher number of examples and changing example selection from random to distance-based. However, none among the various strategies adopted to generate GenAI estimators (in terms of prompting, sample size, example selection strategy) is able to assert itself as superior to the others. 

\subsection{General Fairness}
In terms of fairness, we obtained a variegated mix of results. In general, we can observe only that the adoption of GEN-AI leads to lower bias overall, as indicated by the lower BRIO risk scores attained by GEN-AI models. The same BRIO risk seems to indicate that the table-based prompts should lead to slightly better fairness. Checking each sensitive feature separately, however, EOD values show that GenAI tends to lower only some of the fairness bias, and this without evident trends.

\section{Conclusions}

In this study, we have compared the effectiveness of traditional credit risk evaluation models with approaches based on GenAI. The results show that, in terms of performance metrics, traditional models consistently outperform GenAI-based approaches. Logistic Regression remains the preferred choice for high-stakes credit risk decisions due to its strong balance of predictive performance and interpretability, making it well-suited for regulated environments. It is only in terms of fairness that GenAI models demonstrate slight improvements with respect to the traditional evaluators, suggesting potential for mitigating disparities related to sensitive variables. 

Our findings, which align with previous research \citep{GBabaei_2024, YLi_2024} and complete it, tend to support the idea that current GenAI implementations do not meet the standards required for effective deployment in credit score modeling. In spite of the implementation methodology and the model adopted, current GenAI techniques are shown to lack the precision, stability, and generalization capabilities necessary for competing with risk evaluators based on traditional techniques.

Nonetheless, it cannot be ruled out that further research and methodological refinements could eventually yield improved results in GenAI performance for this task.
More in general, other future research should explore hybrid modeling strategies to unlock the full potential of GenAI in credit risk analytics:

GenAI models hold significant promise in complementary applications within credit risk management. In particular, their ability to extract and synthesize information from unstructured documents—such as loan applications, financial statements, and customer communications—can enhance traditional workflows by generating structured features that feed into classical models like logistic regression. This hybrid approach leverages the strengths of GenAI in natural language understanding and data extraction, while relying on the proven robustness and interpretability of established predictive models. As such, GenAI can serve as a powerful preprocessing tool, improving data quality and enriching feature sets, which may ultimately lead to better risk assessment without replacing existing scoring methodologies.

In conclusion, while modern GenAI models constitute an extremely powerful tool, as of now, they do not seem ready to replace established methods in credit score modeling.In fact, existing approaches based on Logistic Regression, combined with effective variable selection techniques, continue to provide more robust and accurate results. Further research and development will be necessary before GenAI technologies can replace traditional AI risk evaluators.


\bibliographystyle{unsrtnat}
\bibliography{references.bib}

\end{document}